\documentclass[11pt,a4paper]{article}

\usepackage{iftex}
\ifPDFTeX
  \usepackage[utf8]{inputenc}
  \usepackage[T1]{fontenc}
  \usepackage{times}
  \DeclareUnicodeCharacter{2013}{--}
  \DeclareUnicodeCharacter{2014}{---}
  \DeclareUnicodeCharacter{2018}{`}
  \DeclareUnicodeCharacter{2019}{'}
  \DeclareUnicodeCharacter{201C}{``}
  \DeclareUnicodeCharacter{201D}{''}
  \DeclareUnicodeCharacter{03A3}{$\Sigma$}
  \DeclareUnicodeCharacter{2192}{$\to$}
  \DeclareUnicodeCharacter{00D7}{$\times$}
\else
  \usepackage{newtxtext}
\fi
\usepackage[margin=0.92in,top=0.72in,bottom=0.82in]{geometry}
\usepackage{hyperref}
\usepackage{url}
\usepackage{booktabs}
\usepackage{graphicx}
\usepackage{amsfonts}
\usepackage{amsmath}
\usepackage{nicefrac}
\usepackage{microtype}
\usepackage[table]{xcolor}
\usepackage[numbers,sort&compress]{natbib}
\usepackage{caption}
\usepackage{fancyhdr}
\usepackage{tikz}
\usepackage{array}
\usepackage{subcaption}
\usepackage{multirow}
\usepackage{enumitem}
\usepackage[most]{tcolorbox}
\usepackage{seqsplit}
\usepackage{rotating}
\usepackage{calc}
\usepackage{fvextra}

\setlist{nosep,leftmargin=*}

\definecolor{aiBlue}{HTML}{246BFE}
\definecolor{aiCyan}{HTML}{00A6D6}
\definecolor{aiOrange}{HTML}{F59E0B}
\definecolor{aiPurple}{HTML}{7A4CC2}
\definecolor{aiGray}{HTML}{666666}
\definecolor{aiLightGray}{HTML}{E6E6E6}
\definecolor{aiBlack}{HTML}{111111}
\definecolor{labBlue}{HTML}{246BFE}
\definecolor{labCyan}{HTML}{3FB7FF}
\definecolor{ustcBlue}{HTML}{0066B3}
\definecolor{seuGold}{HTML}{F0C400}
\definecolor{seuOrange}{HTML}{D98C00}
\definecolor{seuGreen}{HTML}{4F7F2A}
\definecolor{seuBrown}{HTML}{2B211D}
\definecolor{alibabaOrange}{HTML}{FF6A00}
\definecolor{iflytekRed}{HTML}{D71920}
\definecolor{titleBlack}{HTML}{111111}
\definecolor{edublue}{RGB}{225,240,255}

\hypersetup{
  colorlinks=true,
  linkcolor=aiBlue,
  citecolor=aiBlue,
  urlcolor=aiBlue,
  pdftitle={Edu-Theater: A Data-Efficient Agent Framework for Scalable Learner Behavior Simulation through Staging Roll-Call},
  pdfauthor={Weibo Gao, Qi Liu, Linan Yue, Zheng Zhang, Yichao Du, Fangzhou Yao, Ao Yu, Zhenya Huang, Shijin Wang}
}

\renewenvironment{abstract}
  {\noindent\begin{minipage}{0.98\textwidth}\small\noindent}
  {\end{minipage}\par\medskip}

\newcommand{\paperSubtitle}{A Data-Efficient Agent Framework for Scalable Learner Behavior Simulation through Staging Roll-Call}
\newcommand{\paperAuthors}{Weibo Gao$^{1,2}$, Qi Liu$^{1,2}$, Linan Yue$^{3}$, Zheng Zhang$^{1,2}$, Yichao Du$^{4}$, Fangzhou Yao$^{1,2}$, Ao Yu$^{1,2}$, Zhenya Huang$^{1,2}$, Shijin Wang$^{5,2}$}
\newcommand{\paperAffils}{$^{1}$University of Science and Technology of China \quad $^{2}$State Key Laboratory of Cognitive Intelligence\\[-0.08em] $^{3}$Southeast University \quad $^{4}$Alibaba Group \quad $^{5}$iFLYTEK Co., Ltd.}
\newcommand{\paperEmail}{\href{mailto:weibogao@mail.ustc.edu.cn}{weibogao@mail.ustc.edu.cn}}
\newcommand{\Description}[1]{}

\tcbuselibrary{breakable,skins}
\newtcolorbox{promptbox}[2][]{%
  enhanced,
  breakable,
  sharp corners,
  boxrule=0.8pt,
  colback=white,
  colframe=black!70,
  left=6pt,right=6pt,top=6pt,bottom=6pt,
  fonttitle=\bfseries,
  title={#2},
  before skip=8pt plus 2pt minus 2pt,
  after skip=8pt plus 2pt minus 2pt,
  parbox=false,
  #1
}

\DefineVerbatimEnvironment{promptcode}{Verbatim}{%
  breaklines=true,
  breakanywhere=true,
  fontsize=\small
}
\newenvironment{teaserfigure}{\par\smallskip\noindent\begin{minipage}{\textwidth}\centering\captionsetup{type=figure}}{\end{minipage}\par\medskip}

\newcommand{\LogoBadge}[3]{%
  \begin{tikzpicture}[baseline=(base.base)]
    \node[
      rounded corners=2.2pt,
      draw=#2,
      line width=0.45pt,
      inner xsep=5.2pt,
      inner ysep=3.2pt,
      text=#2,
      font=\scriptsize\bfseries
    ] (base) {#1};
  \end{tikzpicture}%
}
\newcommand{\LABLogo}{%
  \IfFileExists{figures/lab_logo.pdf}{\includegraphics[height=0.52cm]{figures/lab_logo.pdf}}{%
  \IfFileExists{figures/lab_logo.png}{\includegraphics[height=0.52cm]{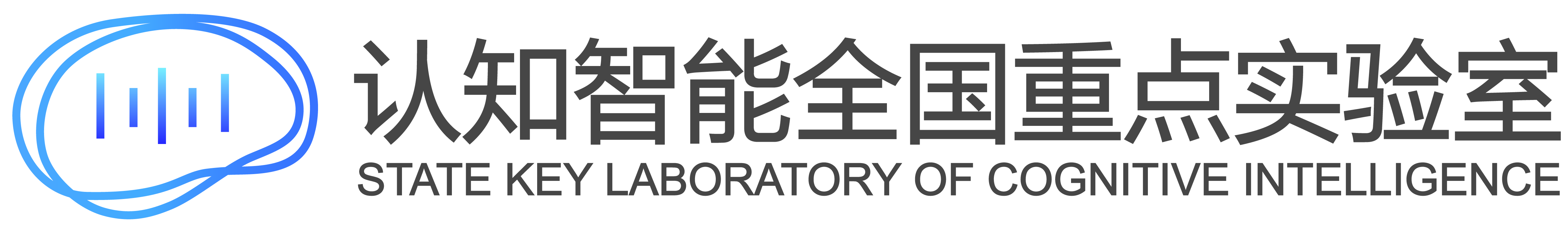}}{\LogoBadge{LAB}{aiBlue}{}}}%
}
\newcommand{\USTCLogo}{%
  \IfFileExists{figures/ustc_logo.pdf}{\includegraphics[height=0.55cm]{figures/ustc_logo.pdf}}{%
  \IfFileExists{figures/ustc_logo.png}{\includegraphics[height=0.55cm]{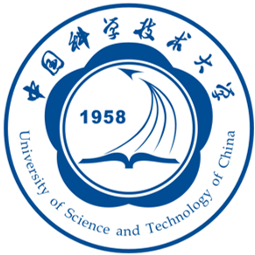}}{\LogoBadge{USTC}{aiBlue}{}}}%
}
\newcommand{\SEULogo}{%
  \IfFileExists{figures/seu_logo.pdf}{\includegraphics[height=0.52cm]{figures/seu_logo.pdf}}{%
  \IfFileExists{figures/seu_logo.png}{\includegraphics[height=0.52cm]{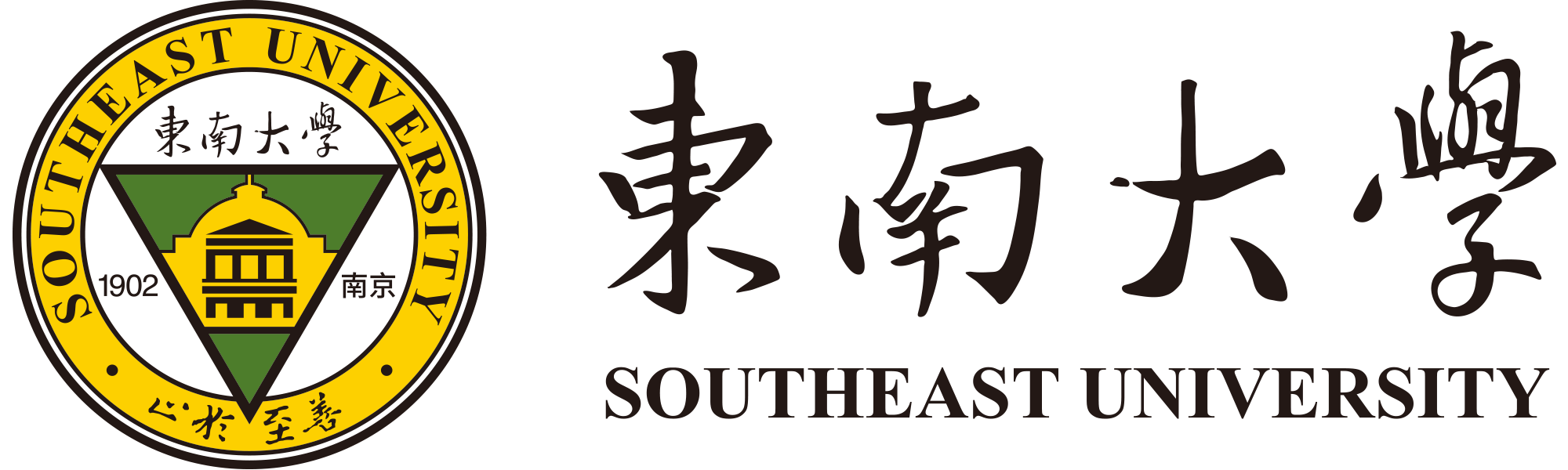}}{\LogoBadge{SEU}{aiPurple}{}}}%
}
\newcommand{\AlibabaLogo}{%
  \IfFileExists{figures/alibaba.pdf}{\includegraphics[height=0.52cm]{figures/alibaba.pdf}}{%
  \IfFileExists{figures/alibaba.png}{\includegraphics[height=0.52cm]{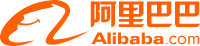}}{\LogoBadge{Alibaba}{alibabaOrange}{}}}%
}
\newcommand{\IFLYTEKLogo}{%
  \IfFileExists{figures/kedaxunfei.pdf}{\includegraphics[height=0.52cm]{figures/kedaxunfei.pdf}}{%
  \IfFileExists{figures/kedaxunfei.png}{\includegraphics[height=0.52cm]{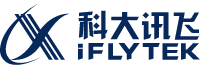}}{\LogoBadge{iFLYTEK}{iflytekRed}{}}}%
}

\newcommand{\EduHeaderWordmark}{%
{\bfseries
\textcolor{labBlue}{E}%
\textcolor{labCyan}{d}%
\textcolor{ustcBlue}{u}%
\textcolor{titleBlack}{-}%
\textcolor{seuGold}{T}%
\textcolor{seuOrange}{h}%
\textcolor{seuGreen}{e}%
\textcolor{seuGold}{a}%
\textcolor{seuOrange}{t}%
\textcolor{seuGreen}{e}%
\textcolor{seuGold}{r}%
\textcolor{titleBlack}{:}%
}}

\newcommand{\EduTitleWordmark}{%
{\LARGE\bfseries
\textcolor{labBlue}{E}%
\textcolor{labCyan}{d}%
\textcolor{ustcBlue}{u}%
\textcolor{titleBlack}{-}%
\textcolor{seuGold}{T}%
\textcolor{seuOrange}{h}%
\textcolor{seuGreen}{e}%
\textcolor{seuGold}{a}%
\textcolor{seuOrange}{t}%
\textcolor{seuGreen}{e}%
\textcolor{seuGold}{r}%
\textcolor{titleBlack}{:}%
}}

\newcommand{\makecustomtitle}{%
  \thispagestyle{empty}%
  \begingroup
  \setlength{\parindent}{0pt}%
  \begin{minipage}[t]{0.73\textwidth}
    \USTCLogo\hspace{0.45em}\LABLogo\hspace{0.45em}\SEULogo\hspace{0.45em}\AlibabaLogo\hspace{0.45em}\IFLYTEKLogo
  \end{minipage}%
  \hfill
  \begin{minipage}[t]{0.23\textwidth}
    \raggedleft\footnotesize\textcolor{aiGray}{Edu-Theater}
  \end{minipage}
  \par\smallskip
  \noindent\textcolor{aiLightGray}{\rule{\textwidth}{0.55pt}}\par
  \medskip
  \EduTitleWordmark\par
  {\LARGE\bfseries \paperSubtitle\par}
  \medskip
  {\small\bfseries \paperAuthors\par}
  \smallskip
  \ifx\paperAffils\empty\else{\footnotesize \paperAffils\par}\fi
  \ifx\paperEmail\empty\else\smallskip{\footnotesize \paperEmail\par}\fi
  \medskip
  \noindent\textcolor{aiLightGray}{\rule{\textwidth}{0.55pt}}\par
  \medskip
  \endgroup
}

\begin{document}

\makecustomtitle

\begin{abstract}
{\Large\bfseries Abstract} \quad
Large-scale learner-task interaction data are crucial for intelligent educational systems but are costly to collect and constrained by privacy and learner engagement. Learner simulators play a critical role in simulating scalable learner behavior without the need for continuous involvement of real learners. However, existing methods are predominantly \textbf{individual-centric}, pairing a simulator with each learner to iteratively infer latent knowledge states from dense interaction histories, which is both data- and computation-intensive, and fragile in cold-start scenarios. We propose a \textbf{cohort-aware roll-call simulation paradigm} that first constructs cohort-level proficiency priors and refines individual learner states through a small number of targeted diagnostic queries. Based on this paradigm, we introduce \textbf{Edu-Theater}, an LLM-powered agent system that performs cohort-aware learner simulation via a teacher agent and retrospective roll-call probing over learner logs. Edu-Theater enables scalable future behavior simulation without the need for dense per-learner histories. Experiments on two real-world datasets demonstrate that Edu-Theater achieves higher simulation accuracy with significantly fewer LLM calls, producing synthetic data that enhances downstream applications such as adaptive testing.
\end{abstract}


\vspace{0.5cm}
\begin{teaserfigure}
  \includegraphics[width=0.98\textwidth]{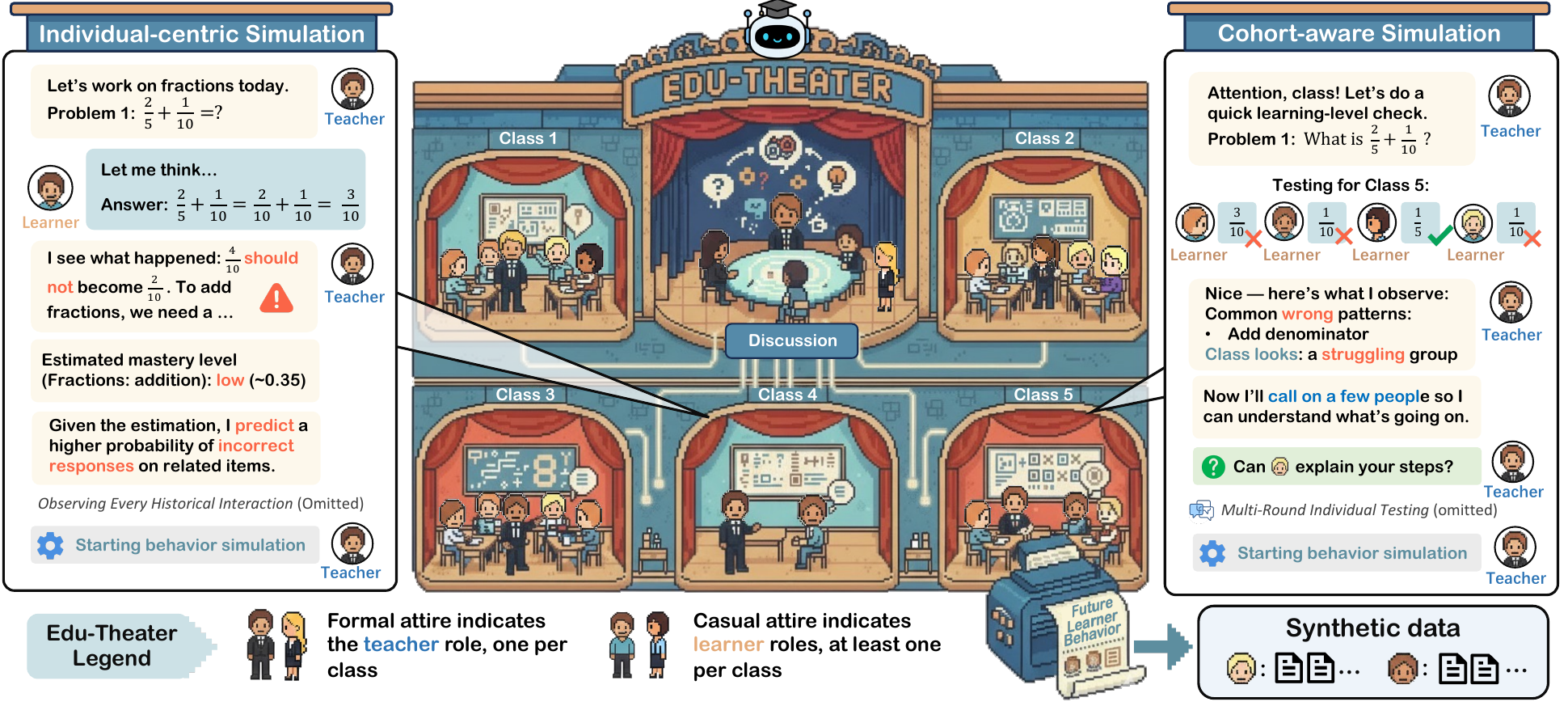}
  \caption{Illustration of Edu-Theater, a learner behavior simulation framework powered by LLM-enabled generative agents. The system simulates a classroom environment where a central Teacher Agent (center) conducts cohort-wide diagnostic probing (right) across learners with diverse cognitive traits and learning histories. Edu-Theater uses population-aware priors to efficiently infer learner states, while also having the flexibility to revert to a traditional `Private Tutor' mode (left) for individualized interactions. Our approach enables scalable, high-fidelity simulations of learner behavior in educational settings, utilizing the power of LLM-based agents to model complex learning processes.}
  \Description{}
  \label{fig:teaser}
\end{teaserfigure}

\section{Introduction}
Education is one of the most interactive domains for generative AI~\cite{dai2024agent4edu}.
Each learning task, such as an exercise or a lesson, elicits learner behaviors and leaves behind interaction traces, including submitted answers and revision trajectories.
Modern tutoring platforms leverage large-scale learner-task interactions to model learning progress and provide personalized support, such as hints and task recommendations~\cite{mladenov2021recsim}.
However, collecting such records is costly because learning requires time and cognitive effort, and data collection is often constrained by privacy policies~\cite{gao2021rcd}.
As a result, training and evaluating personalized support algorithms directly on real learners' interaction data at scale is difficult.
Generative learner simulators therefore serve as a key enabler, producing scalable interaction data from observed histories and supporting efficient offline training and evaluation without continuously involving real learners~\cite{xu2024eduagent}.

We revisit learner simulation from a simple yet fundamental perspective: \emph{How can a simulator infer a learner’s latent state from observed interaction data and use it as evidence to simulate future learning behaviors?}
Most existing approaches~\cite{lu2024generative,xu2024eduagent,gao2025agent4edu,zhan2025coderagent} follow an \textbf{individual-centric} paradigm.
To understand a learner, the simulator is paired \emph{one-to-one} with the same learner over time.
It observes learning traces as the learner engages with learning tasks and gradually forms an individualized estimate of the learner’s knowledge mastery state to predict future behaviors.
In this sense, a learner simulator resembles a companion \textbf{private tutor}.
Its operation can be viewed as presenting learning tasks to the learner, summarizing knowledge mastery from the returned learning traces, and using these signals to simulate future behaviors, as illustrated in the left part of Figure~\ref{fig:teaser}.
This private-tutor-style paradigm relies on sufficiently rich interaction data to enable fine-grained personalization and long-term state tracking, but it comes with a cost.
Dependence on dense histories and repeated model invocations makes it data- and computation-intensive at scale.
Moreover, when observations are limited, individual-level state estimation can become unstable, making long-horizon simulation difficult.
This raises a key question: \emph{Can we make learner simulation more data-efficient, reducing cost while remaining reliable under limited observations?}

To address this challenge, the private-tutoring lens naturally leads us to a contrasting yet familiar classroom setting.
When facing a newly enrolled class, teachers rarely build a deep understanding for every student from the start.
Instead, they often rely on \textbf{roll-call-style diagnostic probing}.
By observing each student’s initial behaviors from only a few learning tasks and drawing on experience with previously taught \emph{similar cohorts}, teachers can quickly form a coarse view of the class’s proficiency distribution without exhaustively querying every student one by one.
During instruction, a few rounds of roll-call questioning are then sufficient to pinpoint each individual’s learning state and anticipate subsequent learning behaviors, since most students differ only slightly from the cohort-level baseline.
This observation suggests that learner simulation can benefit from \textbf{cohort-aware inference}~\cite{bi2023beta}.
The key intuition is that learners are not independent.
Students with similar prior experiences tend to follow similar learning trajectories, and group-level regularities can serve as informative priors for estimating individual latent states.
Motivated by this, we formalize a \textbf{roll-call simulation paradigm}.
The simulator first constructs population-level priors from observed cohort experience by grouping learners with similar histories and estimating cohort-level proficiency distributions.
It then refines each learner’s state through a small number of targeted interactive queries, thereby generating future behaviors while preserving individual variation in a single pass.

Based on this insight, we propose \textbf{Edu-Theater}, a data-efficient, LLM-powered agent system that operationalizes {roll-call-style} learner understanding for scalable learner behavior simulation.
True to its name, each simulation run resembles a staged ``performance'' that combines cohort-aware diagnosis with targeted roll-call probing.
All interactions are retrospective, and no real learners are queried, as illustrated in a toy example in the right part of Figure~\ref{fig:teaser}.
Edu-Theater begins by casting learners into cohorts (``classes'') according to their learning histories and appoints a \emph{teacher agent} to orchestrate the simulation.
Rather than modeling each learner as a full-fledged agent as in previous methods, Edu-Theater represents \emph{learners} as lightweight, indexed log \emph{databases} that return grounded interaction traces when queried.
This design provides reliable evidence while avoiding expensive learner-side long-horizon reasoning.
During rehearsal, the teacher agent maintains a cohort-level knowledge proficiency state and derives individualized learner knowledge proficiency estimates from retrospective roll-call traces, while a cognitive diagnosis prop provides auxiliary mastery references.
When the two mastery assessments are inconsistent, a teacher-side retriever is used to propose informative probing tasks for further refinement.
During this process, teachers from different cohorts can also discuss and share their cohort experiences and, when appropriate, adjust a learner to a more suitable cohort.
Once the assessments align, the learner is considered fully understood.
After rehearsal, the teacher performs scalable future behavior simulation by replaying and re-organizing existing evidence.
Overall, Edu-Theater reframes learner simulation from isolated one-to-one tutoring into a cohort-aware roll-call protocol.
This enables reliable simulation with substantially reduced reliance on extensive interaction histories.
It is worth noting that Edu-Theater maintains the flexibility to switch to a traditional `Private Tutor' mode when each cohort contains a single learner.

Our contributions are summarized as follows:
\begin{itemize}[leftmargin=10pt]
    \item \textbf{New Perspective.}
    We approach learner simulation from a teacher's perspective, emphasizing the importance of data-efficient learner simulation.
    \item \textbf{New Framework.}  
    We introduce Edu-Theater, a cohort-aware, LLM-powered simulation framework that uses population-level priors and individualized probing to efficiently predict learner behaviors with minimal data.
    \item \textbf{Empirical Evaluation.}  We evaluate Edu-Theater through extensive experiments on two real-world datasets, demonstrating its superior performance and cost-efficiency compared to existing simulators, particularly under data-sparse conditions.
\end{itemize}

\section{Preliminary}
\subsection{Roles and Props in Edu-Theater}
\label{sec:roles}

Edu-Theater formulates learner simulation as a staged classroom roll-call, where all interactions are retrospective and no real learners are queried. The system replays historical learner-task interactions, with a teacher agent performing reasoning and generation. Each simulation within a cohort involves two roles: a teacher agent and a cohort of learners, supported by two props for task retrieval and cognitive diagnosis estimation.

\subsubsection{\textbf{Roles in Edu-Theater}}
\paragraph{\textbf{Teacher (Agent) $\mathcal{A}^{\mathcal{T}}$.}}
The teacher agent $\mathcal{A}^{\mathcal{T}}$ is the central actor in Edu-Theater that generates simulated learner behaviors.
It maintains an internal memory state $\mathcal{M}$ summarizing retrieved interaction evidence.
Given a target task $e^\star$ and learner-task interaction traces $\mathcal{Q}_u$, the teacher produces a predicted outcome of learner $u$ as
\begin{equation}
\hat{y}_{u,e^\star}
=
\mathcal{A}^{\mathcal{T}}\!\left(
x_{e^\star}, \mathcal{M}, \mathcal{Q}_u
\right),
\label{eq:teacher_simulate_pd}
\end{equation}
where $\mathcal{Q}_u$ denotes a small set of historical interaction records for learner $u$, with each record in the form of a task--response pair $(x_e, y_{u,e})$, e.g., $x_e$ is a problem statement or lesson prompt, and $y_{u,e}$ is the corresponding answer or action.

\paragraph{\textbf{Learner (Log Database) $\mathcal{D}_u$.}}
Each learner $u$ is represented as a static, indexed log database $\mathcal{D}_u$ that stores historical interactions and serves as the sole evidence source for retrospective roll-call.
Given a task $e$, it returns the corresponding recorded interaction:
\begin{equation}
(x_{e}, y_{u,e}) = \mathcal{D}_u(e).
\label{eq:logdb_if}
\end{equation}

\subsubsection{\textbf{Props in Edu-Theater}}

\paragraph{\textbf{Cognitive Diagnosis Model.}}
The cognitive diagnosis prop is a supervised model that fits collected learner--task interactions and outputs concept-wise mastery estimates for each learner:
\begin{equation}
\boldsymbol{\mu}_u = \mathrm{CD}_{\phi}(\mathcal{Q}_u), \qquad \mu_u(c)\in[0,1].
\label{eq:prop_if}
\end{equation}
The model is shared across all cohorts and trained by fitting task-solving correctness, minimizing the supervised loss $\sum_{u}\mathcal{L}_{\mathrm{CD}}(\phi;\mathcal{Q}_u)$.
In our implementation, we adopt NeuralCD~\cite{wang2020neural} as the cognitive diagnosis prop, where each dimension of $\boldsymbol{\mu}_u$ corresponds to the mastery estimate $\mu_u(c)$ of a knowledge concept $c$.

\paragraph{\textbf{Teacher-side Task Retriever.}}
The task retriever $\mathcal{R}$ proposes a small set of informative probing tasks for each learner, based on the teacher’s current assessment and auxiliary diagnostic references.
The specific retrieval mechanism is detailed in Section~\ref{sec:model}.

\subsection{Problem Formulation}
\label{sec:task}

We consider a learner population $u \in \mathcal{U}$ and a set of learning tasks $e \in \mathcal{E}$.
Each learner is associated with a set of historical interaction logs stored in a learner-specific database $\mathcal{D}_u$.
Given the defined roles and supporting props, Edu-Theater operates in a staged classroom setting, where all interactions are retrospective and no real learners are queried.
Given a future unseen task $e^\star$, the goal of learner simulation is to predict the learner’s response $\hat{y}_{u,e^\star}$ (Eq.~\eqref{eq:teacher_simulate_pd}), while relying only on limited retrospective interaction records and efficient teacher-side model invocations.

\section{Edu-Theater Framework}
\label{sec:model}

\begin{figure*}[t]
	\centering
	\scalebox{0.483}
	{\includegraphics{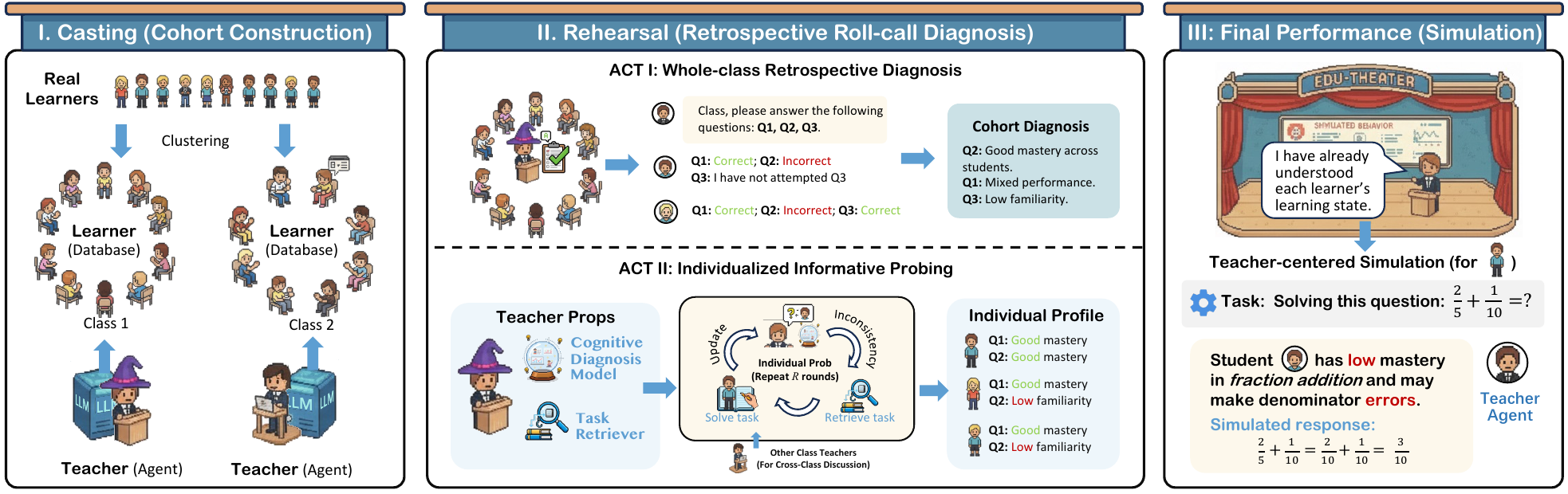}}
	\caption{Overview of the Edu-Theater framework. The simulation process consists of three key phases: \textbf{I. Casting} (cohort construction), \textbf{II. Rehearsal} (retrospective roll-call diagnosis and learner state estimation), and \textbf{III. Final Performance} (teacher-centered scalable simulation). Each learner is represented as a static log database, and the teacher agent orchestrates interaction log retrieval, learner state estimation, and behavior generation.}
	\label{fig:model}
\end{figure*}

Edu-Theater is an LLM-powered agent framework for \emph{roll-call-style} learner simulation with \emph{population-aware inference}.
The framework adopts a theatrical abstraction: the teacher agent conducts a staged classroom performance, while learners are passive log entities whose past behaviors are replayed on demand.
All probing operations are strictly retrospective, and no online interaction with real learners is involved.
The system only reorganizes and summarizes existing historical logs, while the teacher performs inference and generation as if conducting a classroom roll-call.
Following this metaphor, each simulation run consists of three phases, as shown in Figure~\ref{fig:model}:  
\textbf{Casting} (cohort construction),
\textbf{Rehearsal} (retrospective roll-call diagnosis and learner state estimation),
and \textbf{Final Performance} (teacher-centered scalable simulation).
Each learner $u$ is represented as a static log database $\mathcal{D}_u$, while a single teacher agent $\mathcal{A}^{\mathcal{T}}$ orchestrates evidence retrieval, learner state estimation, and behavior generation across all cohorts.
\textbf{All the prompts in this section are provided in Appendix.}

\subsection{Stage I: Casting (Cohort Construction)}
\label{sec:casting}

Casting constructs \emph{cohort-level priors} by grouping learners with similar observed histories.
Each cohort serves as a minimal \emph{classroom unit} that contains at least one learner and is
paired with a dedicated teacher agent.
This design allows the teacher to reuse population-level regularities as informative baselines,
rather than building fully personalized models for every learner from scratch.

For each learner $u$, we compute a representation vector $\mathbf{r}_u$ and cluster
$\{\mathbf{r}_u\}$ into $K$ cohorts $\{C_k\}_{k=1}^{K}$.
Each cohort $C_k$ is associated with a cohort memory $\mathcal{M}_k$ and a cohort-specific
teacher agent $\mathcal{A}^{\mathcal{T}}_k$.
All teacher agents share the same underlying reasoning parameters, but maintain independent
memory states to support population-aware and context-sensitive inference.

When task-solving correctness logs are available, we instantiate $\mathbf{r}_u$ using a
lightweight task-wise status encoding inspired by prior work~\cite{piech2015deep}.
For each task $e\in\mathcal{E}$, we define
\begin{equation}
\mathbf{r}_{u,e} =
\begin{cases}
[1,0,0] & \text{if $u$ attempted $e$ and } y_{u,e}=0,\\
[0,0,1] & \text{if $u$ attempted $e$ and } y_{u,e}=1,\\
[0,1,0] & \text{if $u$ did not attempt $e$},
\end{cases}
\end{equation}
and construct
\begin{equation}
\mathbf{r}_u=\mathrm{Concat}(\{\mathbf{r}_{u,e}\}_{e\in\mathcal{E}})\in\mathbb{R}^{3|\mathcal{E}|}.
\end{equation}
This encoding yields a high-dimensional but sparse representation of each learner’s historical
status over the task space.

When correctness logs are unavailable, $\mathbf{r}_u$ can alternatively be obtained by embedding
an LLM-generated summary of the learner’s historical interactions, depending on specific task requirements.
Casting provides a coarse cohort prior using lightweight features, while fine-grained
learner-specific adaptation is deferred to Stage~II rehearsal.
Cohort assignment may optionally be revised during rehearsal when newly retrieved evidence
indicates systematic mismatch; unless otherwise specified, cohort membership is kept fixed
for stability.
\subsection{Stage II: Rehearsal (Retrospective Roll-call Diagnosis)}
\label{sec:rehearsal}

Rehearsal equips the teacher with a compact yet diagnostic view of each learner by selectively retrieving a small set of historical interaction traces.
Instead of consuming full learner histories or executing long simulated trajectories as in previous methods, Edu-Theater performs \emph{retrospective roll-call} by querying static log databases.
Within each cohort $C_k$, rehearsal proceeds in two acts:
whole-class retrospective diagnosis for cohort calibration, followed by individualized informative probing for learner state estimation refinement.

\subsubsection{\textbf{Act I: Whole-class Retrospective Diagnosis}}
\label{sec:act1}

Act~I aims to obtain a high-coverage cohort sketch by retrieving a diverse set of historical interactions, enabling the teacher to characterize cohort-level proficiency patterns and calibrate the cognitive diagnosis model.

Following coverage-based designs~\cite{bi2020quality}, a probe set $\mathcal{E}^{\mathrm{cov}}$ is constructed to maximize knowledge concept coverage.
Each task $e$ is associated with a knowledge concept set $\mathcal{K}(e)$.
For each concept $c$, we randomly sample $p$ tasks satisfying $c \in \mathcal{K}(e)$
and take their union to construct the probe set:
\begin{equation}
\mathcal{E}^{\mathrm{cov}} = \mathrm{Cover}(\mathcal{E}),
\end{equation}
where $\mathrm{Cover}(\cdot)$ samples $m$ tasks per concept whenever available, and $|\mathcal{E}^{\mathrm{cov}}| \ll |\mathcal{E}|$.

For each learner $u\in C_k$ and each task $e\in\mathcal{E}^{\mathrm{cov}}$, the teacher retrieves the recorded interaction $(x_e,y_{u,e})$ from $\mathcal{D}_u$ if it exists.
Missing records yield no observation.
The retrieved interactions for learner $u$ form a retrospective query set
\begin{equation}
\mathcal{Q}_u=\{(x_e,y_{u,e}) \mid e\in\mathcal{E}^{\mathrm{cov}},\ (x_e,y_{u,e})\in\mathcal{D}_u\}.
\end{equation}

The agent leverages its internal LLM (e.g., GPT-4o) to aggregate the collected cohort-level interaction data into the cohort memory as:
\begin{equation}
\mathcal{M}_k = \mathrm{LLM}\big(\{\mathcal{Q}_u\}_{u\in C_k}\big).
\end{equation}
Here, $\mathcal{M}_k$ stores a compact description of the cohort’s overall knowledge mastery patterns.
The cognitive diagnosis model is refined using aggregated learner--task interaction data $\{\mathcal{Q}_u\}_{u\in \mathcal{U}}$ via gradient-based updates:
\begin{equation}
\phi \leftarrow \phi - \eta \nabla_{\phi} \sum_{u \in \mathcal{U}} \mathcal{L}_{\mathrm{CD}}(\phi; \mathcal{Q}_u),
\end{equation}
where $\eta$ is the learning rate.
At this stage, both the teacher agent and the cognitive diagnosis prop only provide coarse and potentially imperfect assessments of individual learners, which motivates further refinement through subsequent individualized probing.

\subsubsection{\textbf{Act II: Individualized Informative Probing}}
\label{sec:act2}

Act~II reduces uncertainty about individual learners by selectively retrieving additional
evidence through roll-call scenarios.
The teacher rely on cohort-level memory and each learner’s accumulated interaction history to assess the learner.
In each round $r\in\{1,\dots,R\}$, the teacher performs at most one roll-call query for each
learner under a strict probing budget $R$.

Given the current cohort memory $\mathcal{M}_k^{(r)}$ at round $r$
($\mathcal{M}_k^{(0)}=\mathcal{M}_k$) and the learner’s interaction record $\mathcal{Q}_u$,
the teacher infers the learner’s current knowledge state.
The interaction record $\mathcal{Q}_u$ serves as the direct individual evidence, while the
cohort memory provides a population-level prior.
We do not construct a separate long-term memory summary for each learner.
Instead, when a learner exhibits systematic deviations or boundary behaviors, such
observations are incorporated into the shared cohort memory during the update step.
The resulting assessment is represented by a teacher-side estimate
$\tilde{\mu}_u^{(r)}(c)$ for each knowledge concept $c$.

Meanwhile, the cognitive diagnosis prop provides an auxiliary mastery reference $\mu_u(c)$
for learner $u$ on concept $c$, based on the \emph{same individual interaction traces}.
At this stage, both the teacher agent and the cognitive diagnosis prop rely on incomplete
evidence, and their assessments of individual learners are inherently biased and potentially
imperfect.
A learner is considered sufficiently understood only when the two assessments are largely
consistent.

The comparison between these two estimates is therefore used to identify concepts on which
the learner’s state remains ambiguous and may benefit from additional evidence.
We define disagreement as
\begin{equation}
D_u^{(r)}(c)=|\mu_u(c)-\tilde{\mu}_u^{(r)}(c)|,
\end{equation}
and quantify the internal ambiguity of the auxiliary diagnosis by
\begin{equation}
U_u^{(r)}(c)=\mu_u(c)\big(1-\mu_u(c)\big).
\end{equation}
This scoring rule prioritizes concepts that are both inconsistent with the teacher’s
assessment and internally ambiguous under the auxiliary reference.

Based on $D_u^{(r)}(c)$ and $U_u^{(r)}(c)$, the teacher-side task retriever proposes a small
set of candidate probing tasks for learner $u$:
\begin{equation}
\mathcal{C}^{(r)}(u)=\mathcal{R}(\tilde{\mu}_u^{(r)}, \mu_u),
\end{equation}
where each candidate task $e \in \mathcal{C}^{(r)}(u)$ is scored by
\begin{equation}
\mathrm{Score}^{(r)}(u,e)=\sum_{c\in\mathcal{K}(e)}
\big(D_u^{(r)}(c)+U_u^{(r)}(c)\big).
\end{equation}
The highest-scoring task $e^{(r)}$ is selected and its historical interaction
$(x_{e^{(r)}}, y_{u,e^{(r)}})$ is retrieved from $\mathcal{D}_u$ if it exists.
If the record is missing, no observation is returned.
The retrieved interaction is appended to
\begin{equation}
\mathcal{Q}_u \leftarrow \mathcal{Q}_u \cup \{(x_{e^{(r)}},y_{u,e^{(r)}})\}.
\end{equation}

After all learners are processed in the current round, the cohort memory is updated:
\begin{equation}
\mathcal{M}_k^{(r+1)}=\mathrm{LLM}\big(\{\mathcal{Q}_u\}_{u\in C_k}\big),
\end{equation}
which revises the teacher’s class-level priors.
The cognitive diagnosis prop is then incrementally refined using the updated
$\{\mathcal{Q}_u\}_{u\in C_k}$ as auxiliary diagnostic references.


After the cohort memory is updated in each round, the teacher agent broadcasts the
profiles of the learners it has just probed to a small subset of other teacher agents,
and queries them one by one about whether these learners would fit better in their
cohorts.
Each receiving teacher compares the learner’s interaction history with its own cohort
summary and returns a binary decision.
Once a cohort that provides a more consistent explanation is found, the learner is
reassigned to that cohort and the cross-cohort review for this learner terminates.
We do not exhaustively query all teachers or search for the globally best cohort.
Although doing so could yield a more optimal reassignment, it would incur prohibitive token costs, and in practice identifying a locally better cohort already provides a meaningful improvement over making no adjustment.

This iterative roll-call process continues until the probing budget is exhausted
($r=R$), indicating that further queries are
unlikely to yield informative gains.

\subsection{Stage III: Final Performance (Teacher-centered Simulation)}
\label{sec:performance}

After rehearsal, the teacher agent is treated as a calibrated performer that encodes both
cohort-level learning patterns and individual learner histories.
Edu-Theater then conducts scalable simulation with the teacher as the only agentic actor,
while learners remain passive log entities.
At this stage, the cognitive diagnosis prop can also be applied to the same interaction
records to provide auxiliary concept-wise mastery estimates $\boldsymbol{\mu}_u$ as a
reference.

Given a target task $e^\star$, the teacher generates a simulated learner response by
conditioning on the task content, the cohort memory, and the learner’s interaction history:
\begin{equation}
\hat{y}_{u,e^\star}
=
\mathcal{A}^{\mathcal{T}}\!\left(
x_{e^\star}, \mathcal{M}_k, \mathcal{Q}_u,\boldsymbol{\mu}_u
\right),
\end{equation}
where $\hat{y}_{u,e^\star}$ is the simulated outcome for learner $u$ on task $e^\star$.

\subsection{Computational Cost}
\label{sec:cost}
The computational cost of Edu-Theater is mainly determined by LLM invocations and token consumption. In Act I, the system performs at most \(|\mathcal{U}| |\mathcal{E}^{\mathrm{cov}}|\) retrospective log queries, where \(|\mathcal{U}|\) is the number of learners and \(|\mathcal{E}^{\mathrm{cov}}|\) is the size of the coverage probe set. In Act II, it conducts at most \(|\mathcal{U}| R\) informative queries under the probing budget \(R\). The total number of queries is thus \(O(|\mathcal{U}| (|\mathcal{E}^{\mathrm{cov}}| + R))\). The number of LLM invocations is \(K(R+1)\): \(K\) initial calls for cohort memory construction and \(KR\) updates during iterative rehearsal. The total log units processed by the LLM are approximately \(|\mathcal{U}| |\mathcal{E}^{\mathrm{cov}}| + R|\mathcal{U}|\). By comparison, traditional agent methods require incremental full-sequence replay, leading to per-learner cost proportional to \(T_{\text{avg}}(T_{\text{avg}}+1)/2\) and overall population cost \(|\mathcal{U}| \cdot T_{\text{avg}}(T_{\text{avg}}+1)/2\), where \(T_{\text{avg}}\) is the average number of tasks per learner. In our experiments, \(R \approx 15{\sim}20\) and \(K = 15\), both small relative to typical sequence lengths. By only accessing a compact, informative subset of logs instead of replaying full histories, Edu-Theater significantly reduces both LLM calls and token usage, yielding a much lighter computational footprint.
More details are provided in Appendix.

\section{Experiments}
\label{sec:exp}

Edu-Theater is designed to enable reliable learner response simulation under \emph{limited retrospective interaction traces} with substantially reduced LLM-side cost.
We evaluate Edu-Theater through the following research questions:
\textbf{RQ1:} Under varying observation availability, how faithfully can Edu-Theater reproduce real learner response patterns?
\textbf{RQ2:} How much teacher-side cost can Edu-Theater reduce compared with individual-centric LLM simulators and classical supervised simulators?
\textbf{RQ3:} Which staged components (Casting, Act~I coverage diagnosis, Act~II discrepancy-driven probing) are necessary for stability and accuracy?
\textbf{RQ4:} Can the simulated responses generated by Edu-Theater improve downstream educational applications such as adaptive testing?

\subsection{Basic Setup}
\paragraph{\textbf{Dataset}}
We evaluate learner simulation on two task-solving datasets: DBE-KT22~\cite{abdelrahman2022dbe} and EduHS.
DBE-KT22 consists of online exercise interactions collected from undergraduate students enrolled in a Relational Databases course at the Australian National University (2018--2021) via the CodeBench platform.
EduHS is collected from a private online practice system and contains interaction logs from high-school learners on mathematics and physics exercises.
Both datasets consist of multiple-choice exercise interactions, where each task corresponds to a specific exercise and each response contains the learner’s selected option and a binary correctness label.
Each record is associated with textual descriptions of the exercise and its annotated knowledge concepts.
For both datasets, we randomly sample 500 learners, each with at least 20 interactions, for experiments.
EduHS will be released upon acceptance.
Table~\ref{tab:statistics} summarizes dataset statistics.

\begin{table}
\centering
\small
\caption{The statistics of two datasets.}
{
{
\begin{tabular}{l|cc}
    \toprule
    Datasets & DBE-KT22 & EduHS \\
    \midrule
    \#learners & 500 & 500 \\

    \#tasks & 183 & 1,032 \\
    
    \#knowledge concepts & 67 & 458 \\


    \#responses & 51,150 & 18,045 \\
    
    \#responses per learners & 102.3 & 36.09 \\

    \bottomrule
\end{tabular}}}
\label{tab:statistics}
\end{table}


\paragraph{\textbf{Baselines.}}
We select two categories of simulators as baselines.

\emph{(i) Supervised simulators.}
We include KES~\cite{liu2019exploiting}, DKVMN~\cite{zhang2017dynamic}, EERNN (with Markov)~\cite{su2018exercise}, SAKT~\cite{pandey2019self}, and DAISIM~\cite{zhao2023simulating} as supervised simulation baselines.
These methods do not rely on LLMs and model learner task-solving outcomes only at the correctness level, formulating simulation as a binary classification problem (predictions above 0.5 are treated as correct).
Among them, EERNN incorporates textual exercise content, whereas the others mainly operate on historical correctness traces.

\emph{(ii) LLM-based generative simulators.}
We include EduAgent~\cite{xu2024eduagent} and Agent4Edu~\cite{gao2025agent4edu} with multiple LLM implementations as generative baselines.
Because expert-annotated cognitive factors are unavailable, we follow prior work by training NeuralCD~\cite{wang2020neural} and using its concept-wise mastery estimates as auxiliary cognitive factors in the prompt.
We also include a \emph{history-conditioned LLM} baseline that directly predicts a learner’s response from a fixed number of retrieved historical records, without cohort memory or auxiliary diagnostic modeling.

\begin{table}[t]
\centering
\caption{Performance on \textbf{DBE-KT22}, with full results on two datasets listed in Appendix.
Higher is better ($\uparrow$). 
``--'' indicates that methods cannot predict answers. 
Best results are in bold. 
Within each backbone LLM, the best-performing agent is highlighted with a light-blue background.}
\label{tab:prediction}

\setlength{\tabcolsep}{2pt}
\renewcommand{\arraystretch}{1}
\small

\begin{tabular}{l|c|ccc}
\toprule
 & \multicolumn{1}{c|}{\textbf{Answer}} & \multicolumn{3}{c}{\textbf{Correctness}} \\
\cmidrule(lr){2-2}\cmidrule(lr){3-5}
\textbf{Model} 
& \textbf{ACC} $\uparrow$
& \textbf{ACC} $\uparrow$
& \textbf{F1-score} $\uparrow$
& \textbf{ROUGE-3}$\uparrow$\\
\midrule

KES & -- & 65.41 & 74.92 & 35.17 \\
DKVMN & -- & 67.79 & 79.50 & 40.84 \\
EERNN & -- & 69.32 & 79.66 & 42.15 \\
SAKT & -- & 68.92 & 81.13 & 34.39 \\
DAISIM & -- & 69.03 & 81.05 & 34.92 \\
\midrule

Llama3-8b & 59.40 & 66.13 & 78.24 & 40.05 \\
GPT-3.5-turbo & 57.60 & 65.05 & 75.21 & 35.77 \\
GPT-4.1-mini & 58.10 & 64.67 & 79.30 & 35.13 \\
\midrule
\multicolumn{5}{c}{\textbf{Agent-based Methods (Grouped by Backbone LLM)}}\\
\midrule

EduAgent (Llama2-7b) & 61.90 & 68.32 & 78.82 & 41.25 \\
Agent4Edu (Llama2-7b) & 63.80 & 69.30 & 79.92 & 42.13 \\
Edu-Theater (Llama2-7b) & 
\cellcolor{edublue}65.20 & \cellcolor{edublue}69.55 & \cellcolor{edublue}80.21 & \cellcolor{edublue}42.12 \\
\cmidrule(lr){1-5}

EduAgent (Llama3-8b) & 62.30 & 68.45 & 79.22 & 41.40 \\
Agent4Edu (Llama3-8b) & 64.10 & 69.50 & 80.62 & 42.25 \\
Edu-Theater (Llama3-8b) & 
\cellcolor{edublue}65.80 & \cellcolor{edublue}69.77 & \cellcolor{edublue}80.88 & \cellcolor{edublue}42.33 \\
\cmidrule(lr){1-5}

EduAgent (GPT-3.5-turbo) & 63.20 & 69.70 & 80.72 & 42.25 \\
Agent4Edu (GPT-3.5-turbo) & 64.80 & 70.30 & 81.82 & 42.57 \\
Edu-Theater (GPT-3.5-turbo) & 
\cellcolor{edublue}67.10 & \cellcolor{edublue}71.72 & \cellcolor{edublue}83.21 & \cellcolor{edublue}43.63 \\
\cmidrule(lr){1-5}

$\text{EduAgent}$ (GPT-4o) & 65.30 & 69.82 & 80.82 & 42.40 \\
$\text{Agent4Edu}$ (GPT-4o) & 66.90 & 70.82 & 82.13 & 42.73 \\
Edu-Theater (GPT-4o) & 
\cellcolor{edublue}68.60 & \cellcolor{edublue}72.93 & \cellcolor{edublue}83.72 & \cellcolor{edublue}\textbf{43.73} \\
\cmidrule(lr){1-5}

$\text{EduAgent}$ (GPT-4.1-mini) & 65.90 & 69.95 & 81.22 & 42.60 \\
$\text{Agent4Edu}$ (GPT-4.1-mini) & 67.40 & 71.30 & 82.93 & 43.15 \\
Edu-Theater (GPT-4.1-mini) & 
\cellcolor{edublue}\textbf{69.40} & \cellcolor{edublue}\textbf{73.03} & \cellcolor{edublue}\textbf{83.95} & \cellcolor{edublue}42.85 \\
\bottomrule
\end{tabular}

\end{table}

\paragraph{\textbf{Implementation}} \label{sec:setup}
We implement the teacher agent using both API-based and open-source LLMs.
For API-based models, we use the OpenAI service\footnote{\url{https://platform.openai.com/docs/models}} and adopt GPT-3.5-turbo-1106 (denoted as GPT-3.5), GPT-4o-2024-11-20 (denoted as GPT-4o), and GPT-4.1-mini.
For open-source models, we use Llama2-7B and Llama3-8B~\cite{touvron2023llama}.
The temperature of all LLMs is fixed at 0 to ensure deterministic generation.
For \textbf{Edu-Theater}, we conducted simulations with various parameter settings (see Section~\ref{sec:rq3} for details). Specifically, we report the results for \( K = 15 \), \( p = 10 \), and \( R = 15 \) on the DBE-KT22 dataset, and \( K = 15 \), \( p = 15 \), and \( R = 20 \) on the EduHS dataset.
To train NeuralCD, we adopt the original parameters from its corresponding paper, with the learning rate $\eta$ set to 0.002.
We report the mean over five runs on a Linux server equipped with 4 $\times$ NVIDIA A100 80GB GPUs.

\paragraph{\textbf{Evaluation Metrics for Simulation Effectiveness.}}
We evaluate simulation effectiveness from two aspects: (1) whether the simulated model selects the same answer option as the real learner, measured by accuracy (ACC); and (2) whether the predicted task-solving correctness matches the ground truth, measured by ACC and F1-score. Following~\cite{zhao2023simulating}, we further adopt ROUGE-3 to assess the similarity between the simulated correctness and the real correctness distribution.

\begin{table}
\centering
\caption{Correctness simulation accuracy on the DBE-KT22 dataset under cold-start scenarios.}
\small
\scalebox{0.95}{
\begin{tabular}{l|ccccc}
\toprule
Model & 0\% & 20\% & 30\% & 50\% & 100\% \\ \midrule

KES & - & 55.31 & 60.13 & 63.27 & 65.41 \\
DKVMN & - & 60.14 & 64.32 & 65.33 & 67.79 \\
EERNN & - & 68.53 & 68.35 & 69.04 & 69.32 \\
SAKT & - & 67.88 & 67.92 & 68.37 & 68.92 \\
DAISIM & - & 65.25 & 66.33 & 68.44 & 69.03 \\ \midrule
EduAgent (Llama3-8b) 
& 63.25 & 65.67 & 65.30 & 67.33 & 68.45 \\
EduAgent (GPT-4.1-mini) 
& 66.93 & 68.42 & 68.57 & 68.52 & 69.95 \\
\midrule
Agent4Edu (Llama3-8b) 
& 65.83 & 66.02 & 68.30 & 69.63 & 69.50 \\ 
Agent4Edu (GPT-4.1-mini)
& 68.43 & 69.32 & 69.43 & 69.95 & 71.30 \\ 
\midrule
Edu-Theater (Llama3-8b)  
& {66.50} & {66.33} & {68.67} & {69.67} & {69.77} \\ 
Edu-Theater (GPT-4.1-mini) 
& \textbf{68.85} & \textbf{69.90} & \textbf{70.33} & \textbf{71.27} & \textbf{73.03} \\ \bottomrule
\end{tabular}}
\label{tab:cold}
\end{table}


\subsection{Simulation Performance (RQ1)}
The goal of Edu-Theater is to generate simulated learner responses that resemble real responses under limited retrospective interaction records.
We therefore evaluate response simulation performance in two settings: a data-available scenario and a cold-start scenario.

\paragraph{\textbf{Data-Available Scenario.}}
In this setting, each learner's interaction log is split into 90\% observed records (for optimization) and 10\% held-out future tasks (for test).
Supervised baseline models are trained on the observed records, with the last 20\% of each learner's observed records used for validation.
Edu-Theater is allowed to retrieve retrospective interaction records during rehearsal (Stage~II) and then generate responses for held-out tasks during final performance (Stage~III).
The learner's task-solving response simulation performance on \textbf{DBE-KT22} is summarized in Table~\ref{tab:prediction}, with \textbf{the full results on \textbf{DBE-KT22} and \textbf{EduHS} listed in Appendix}.

The results indicate that:
(1) LLM-based simulators consistently outperform supervised baselines and native LLMs, demonstrating a stronger ability to simulate learner task-solving behavior.
(2) Among supervised models, EERNN performs best due to its use of exercise text.
(3) GPT-based teachers outperform Llama-based teachers, suggesting stronger capacity in modeling learner–task interaction patterns.
(4) All models achieve higher performance on DBE-KT22 than on EduHS, as DBE-KT22 contains denser learner interaction logs.
(5) The answer-level simulation accuracy is consistently lower than correctness-level performance, since answer prediction involves a larger output space and thus poses a more challenging learning problem.

\paragraph{\textbf{Cold-Start Scenario.}}
In the cold-start setting, we construct five levels of data sparsity by randomly selecting 0\%, 20\%, 30\%, 50\%, and 100\% of each learner's observed records.
Supervised models are trained on the available subset, while Edu-Theater initializes cohort construction and rehearsal using only the same limited records.

Results on DBE-KT22 are reported in Table~\ref{tab:cold}, and similar trends are observed on the EduHS dataset.
We observe that:
(1) When no historical records are available (0\%), supervised models cannot operate, while Edu-Theater remains functional.
(2) As observation becomes sparser, supervised models degrade sharply, whereas Edu-Theater maintains stable performance.
(3) Edu-Theater achieves the highest robustness under cold-start conditions, benefiting from cohort-level priors and population-aware inference.

\subsection{LLM-based Learner Simulation Cost (RQ2)}
LLM-based simulators generally incur higher computational costs than supervised models
due to the repeated invocation of large language models during simulation.
To demonstrate that Edu-Theater reduces such costs compared with existing agents,
we analyze both data usage and API computational expense.

\paragraph{\textbf{Data Cost}}
A key advantage of Edu-Theater lies in its data efficiency during the casting and rehearsal stages.
Instead of requiring full historical interaction logs to initialize learner states,
Edu-Theater constructs cohort-level priors using only a subset of retrospective interaction records,
and refines individual learners through a small number of roll-call queries.
Specifically,
Edu-Theater initializes using only 63.4\% of the DBE-KT22 data
and 69.5\% of the EduHS data,
while baseline LLM-based simulators rely on the full training dataset.
This property makes Edu-Theater particularly suitable for cold-start settings
and scenarios where learner histories are incomplete or sparsely observed.

\begin{table}[t]
    \centering
    \caption{Time and monetary costs of GPT-based LLM simulators.
    $\downarrow$ indicates lower is better.}
    \label{tab:cost}
    \small
    \setlength{\tabcolsep}{2.5pt}
    \renewcommand{\arraystretch}{1}
    \begin{tabular}{l|cc|cc} \hline 
    \multirow{2}{*}{Model}  & \multicolumn{2}{c|}{DBE-KT22} & \multicolumn{2}{c}{EduHS}\\
    \cline{2-5}
    & Time $\downarrow$ & Money $\downarrow$ & Time $\downarrow$ & Money $\downarrow$ \\ \hline 
        EduAgent (GPT-3.5-turbo) & 4h 27m & 47.5\$ & 3h 35m & 27.4\$ \\
        Agent4Edu (GPT-3.5-turbo) & 6h 30m & 62.7\$ & 5h 20m & 43.5\$ \\
        Our Edu-Theater (GPT-3.5-turbo) & \textbf{3h 52m} & \textbf{42.3\$} & \textbf{2h 24m} & \textbf{20.1\$} \\ \hline

        EduAgent (GPT-4.1-mini) & 4h 53m & 44.7\$ & 3h 25m & 25.3\$ \\
        Agent4Edu (GPT-4.1-mini) & 6h 55m & 59.0\$ & 5h 33m & 40.5\$ \\
        Our Edu-Theater (GPT-4.1-mini) & \textbf{4h 24m} & \textbf{39.3\$} & \textbf{2h 23m} & \textbf{18.5\$} \\ \hline
    \end{tabular}
\end{table}

\paragraph{\textbf{Time \& Monetary Cost}}
Table~\ref{tab:cost} reports the total wall-clock time and monetary costs
for representative LLM-based simulators to complete both the rehearsal phase
(retrospective roll-call diagnosis) and the final performance phase
(teacher-centered response generation), under identical experimental settings.
All compared agents follow the same simulation pipeline and API usage protocol.
The reported costs include all API calls made during a full simulation run.
As shown, Edu-Theater consistently requires less computational time
and lower monetary expense than existing LLM-based methods,
demonstrating its efficiency in both learner state estimation
and scalable response generation.
A similar trend is observed on EduHS.

\subsection{Parameter and Module Analysis (RQ3)}

\begin{figure}[t]
	\centering
	\scalebox{0.34}
	{\includegraphics{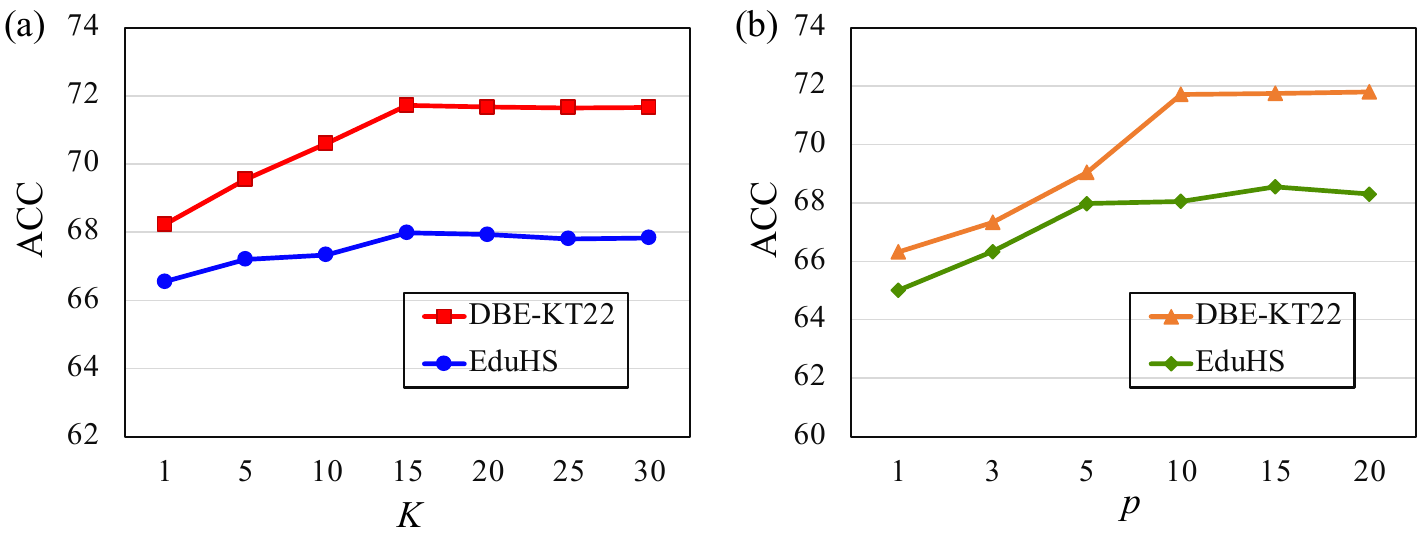}}
	\caption{(a) Performance under different cohort numbers $K$. 
        (b) The impact of whole-class diagnosis density ($p$).}
	\label{fig:fig1}
\end{figure}

\paragraph{\textbf{Cohort Number ($K$) Analysis}} 
\label{sec:rq3}
We analyze how the number of cohorts affects the simulation performance of Edu-Theater.
Figure~\ref{fig:fig1}~(a) shows the accuracy of the GPT-3.5-turbo-based Edu-Theater
with $K$ set to 1, 5, 10, 15, 20, 25 and 35.
As $K$ increases, performance improves and stabilizes after $K = 15$.
Finer cohort construction leads to more homogeneous learning patterns within each cohort,
which provides more informative population-level priors for roll-call simulation.
When $K = 1$, all learners are grouped into a single cohort,
resulting in degraded performance due to the high diversity of learner behaviors.

\paragraph{\textbf{Whole-class Diagnosis Density Analysis ($p$)}}
We analyze how the number of sampled tasks per knowledge concept at the Whole-class Diagnosis stage affects simulation performance.
Recall that in Act~I of rehearsal,
for each concept $c$ we sample $p$ tasks satisfying $c \in \mathcal{K}(e)$ to construct the probe set $\mathcal{E}^{\mathrm{cov}}$.
We vary $p \in \{1, 3, 5, 10, 15, 20\}$ in the experiments.
Figure~\ref{fig:fig1}~(b) shows a clear performance gain as $p$ increases from 1 to 5 on EduHS and from 1 to 10 on DBE-KT22, indicating that denser concept coverage allows the teacher to form more accurate cohort-level proficiency sketches.
However, the improvement becomes marginal when $p$ exceeds 10,
and almost saturates at $p=10 \sim 20$.
This suggests that a small number of representative tasks per concept
is already sufficient for effective cohort calibration,
and additional retrospective queries mainly introduce redundant information.
Overall, the analysis demonstrates that Edu-Theater can achieve strong
simulation performance with relatively low rehearsal cost,
highlighting the efficiency of its coverage-based roll-call design.

\begin{figure}[t]
	\centering
	\scalebox{0.3}
	{\includegraphics{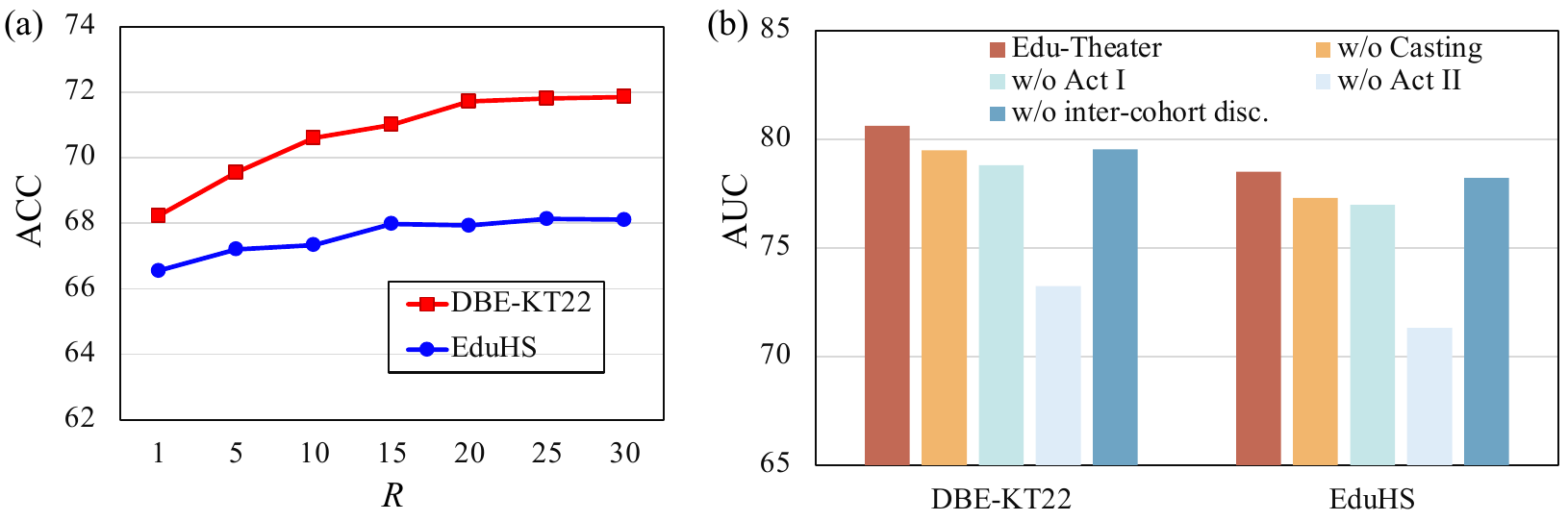}}
	\caption{(a)  The impact of Budget ($R$). (b) Ablation Study.}
	\label{fig:fig2}
 \end{figure}

\paragraph{\textbf{Individualized Informative Probing Budget ($R$)}} 
We investigate how the probing budget $R$ of individualized roll-call
affects simulation performance.
Recall that in Act~II of rehearsal,
the teacher performs at most $R$ retrospective queries for each learner.
We vary $R \in \{1, 5, 10, 15, 20, 25, 30\}$ for GPT-4.1-mini-based Edu-Theater.
As shown in Figure~\ref{fig:fig2}~(a), increasing \( R \) consistently improves performance on EduHS, with gains gradually stabilizing when \( R \geq 20 \) on DBE-KT22 and \( R \geq 15 \) on EduHS.
Longer roll-call sequences provide richer retrospective interaction evidence, leading to more accurate learner state estimation.

\paragraph{\textbf{Ablation Study}}
We conduct an ablation study by removing key stages of the Edu-Theater
roll-call framework to examine their individual contributions.
Specifically, we consider the following variants:
(i) \emph{w/o Casting}, where cohort construction is removed and all learners
share a single teacher memory;
(ii) \emph{w/o Act I}, where cohort-level retrospective diagnosis is skipped
and the teacher directly performs individualized probing;
(iii) \emph{w/o Act II}, where individualized informative probing is disabled
and learner states rely solely on cohort-level summaries;
and (iv) \emph{w/o inter-cohort disc.}, where teacher agents do not discuss and exchange cohort memories during rehearsal.
As shown in Figure~\ref{fig:fig2}~(b), removing any component leads to a clear degradation in simulation performance. Eliminating Casting or Act~II causes the most significant drops, emphasizing the importance of learner-specific refinement beyond coarse cohort estimates. Removing Act~I also results in a substantial performance loss, highlighting the importance of population-level priors and cohort calibration for a reliable diagnostic context. Disabling inter-cohort discussion only slightly decreases performance, suggesting it provides additional constraints for boundary learners, but with less impact than other components. Overall, the ablation results confirm that Edu-Theater’s effectiveness stems from the integration of cohort construction, cohort-level rehearsal, and individualized roll-call refinement, rather than any single component.

\subsection{Education Algorithm Improvement (RQ4)}
We investigate whether the simulated data generated by Edu-Theater can improve downstream intelligent education algorithms.
We choose Computerized Adaptive Testing (CAT) as the evaluation task, as it is a representative and widely used paradigm in personalized assessment.
CAT relies on a cognitive diagnostic model (e.g., IRT~\cite{baker2001basics}) to estimate learner abilities and designs an item selection strategy
to adaptively recommend informative questions.
The collected responses are then used to refine the diagnostic model and predict future learner performance.
If the simulated data generated by Edu-Theater can enhance the prediction accuracy of the diagnostic model, this would indicate that the simulated responses
provide useful and realistic learning signals.

\begin{table}[t]
    \centering
    \small 
    \setlength{\tabcolsep}{2.5pt} 
    \renewcommand{\arraystretch}{1} 
    \caption{The improvement of CAT services on DBE-KT22.}
    \begin{tabular}{l|ccc|ccc} \hline 
    & \multicolumn{3}{c|}{Testing length is 5} & \multicolumn{3}{c} {Testing length is 10}\\
    \toprule
         Model & F1-score $\uparrow$ &  F1-score+ $\uparrow$ & Imp. $\uparrow$ &  F1-score $\uparrow$  & F1-score+ $\uparrow$ & Imp. $\uparrow$ \\ \midrule 
         FSI & 83.33 & 84.87 & +1.54 & 85.16 & 86.17 & +1.01 \\
        KLI & 81.86 & 85.13 & +3.27 & 84.30 & 85.69 & +1.39 \\
        MAAT & 82.59 & 83.90 & +1.31 & 82.87 & 83.70 & +0.83 \\
        \bottomrule
    \end{tabular}
    
    \label{tab:improvement}
\end{table}
We use 60\% of learner interaction data
from DBE-KT22 to train the IRT model,
and reserve the remaining 40\% for evaluation.
For each learner in the test set,
Edu-Theater simulates responses to 20 randomly selected unseen exercises
based on the calibrated teacher and learner evidence.
The generated responses are combined with the original training set
to construct an augmented dataset, denoted as DBE-KT22+.
We then retrain the IRT model using both DBE-KT22 and DBE-KT22+
under three standard CAT strategies,
including FSI~\cite{lord2012applications},
KLI~\cite{chang1996global}, and MAAT~\cite{bi2020quality}.
Each strategy recommends either 5 or 10 test items per learner.
Table~\ref{tab:improvement} reports the IRT prediction performance,
where F1-score denotes performance on the original dataset,
and F1-score+ denotes performance after augmentation.
The results show consistent improvements across all strategies.
This suggests that Edu- Theater can generate high-quality learner response data for the applications.
\section{Related Work} \label{app:related_work}
\subsection{Learner Simulation}
Learner simulation addresses the scarcity of high-quality interaction data in educational systems~\cite{zhao2023simulating, yao2024adard}. Early methods primarily trained classifiers on historical data to predict future responses~\cite{liu2019exploiting, zhao2023simulating}, employing memory-based architectures~\cite{reddy2017accelerating} or RNN variants such as EERNN~\cite{su2018exercise} and KES~\cite{liu2019exploiting}. While DAISim~\cite{zhao2023simulating} utilizes Markov decision processes to capture multi-scale patterns, such traditional methods often struggle with interpretability and cold-start challenges. Recently, LLM-based generative agents have emerged as a robust alternative. Specifically, GenStu~\cite{lu2024generative} generates exercise responses, EduAgent~\cite{xu2024eduagent} simulates instructional learning, and Agent4Edu~\cite{gao2025agent4edu} incorporates forgetting dynamics across interaction turns. Furthermore, CoderAgent~\cite{zhan2025coderagent} extends these simulations to the coding task. Despite their potential, current LLM agents for individual learner simulation entail high computational costs and rely heavily on the model's inherent generalization capabilities. 
In this work, we introduce \textbf{Edu-Theater}, a novel framework that reduces computational costs and maintains effective learner simulation by leveraging cohort-level priors and retrospective roll-call probing, making it ideal for data-scarce scenarios.

\subsection{LLM-based Agents}
LLM-based generative agents have demonstrated exceptional capabilities in perception, decision-making, and execution, sparking extensive cross-domain research~\cite{wang2024survey}. A foundational architecture proposed by Park et al.~\cite{park2023generative}—integrating profiles, memory, action, and reflection—has enabled the simulation of complex, temporally extended human behaviors. Building on this, subsequent research has specialized agents for communication, tool usage, recommendation, and interactive decision-making~\cite{qian2023communicative, wu2023autogen, wang2023voyager, huang2023recommender, zhang2023agentcf, zhang2024generative}, alongside large-scale multi-agent simulations~\cite{gao2023s, wang2023user, liu2023training, wang2023recagent}. In educational contexts, LLM-based agents have been increasingly leveraged to support teaching and classroom interactions~\cite{li2024bringing, dan2023educhat, kieser2023educational, baidoo2023education}. Beyond evaluating ChatGPT's utility in specialized fields like engineering~\cite{qadir2023engineering, rahman2023chatgpt}, recent work emphasizes pedagogical alignment. 
For instance, 
EduAgent~\cite{xu2024eduagent} and Agent4Edu~\cite{gao2025agent4edu} instantiate learner simulators by encoding personalized profiles and workflows, while SimClass~\cite{zhang2024simulating} extends this to classroom-level activities through role-based agent modeling.

\section{Conclusion}
In this work, we proposed a novel approach to learner simulation through \textbf{Edu-Theater}, a data-efficient, cohort-aware simulation framework powered by large language models (LLMs). By leveraging population-level priors and targeted roll-call-style diagnostic probing, Edu-Theater offers a significant departure from traditional one-to-one tutoring paradigms, making learner behavior simulation more scalable and efficient.
Edu-Theater demonstrated the feasibility of achieving personalized and accurate learner simulations with minimal data, even under sparse observation conditions. Through empirical evaluation, we showed that our approach outperforms existing simulators in both cost-effectiveness and prediction accuracy.
By reframing learner simulation from individual-centric models to a cohort-aware, roll-call protocol, we paved the way for more efficient, cost-effective, and scalable simulations that can support the next generation of intelligent educational systems. Future work could explore further optimization of cohort clustering strategies and the incorporation of additional learner attributes for even finer-grained personalization.
\bibliographystyle{ACM-Reference-Format}
\bibliography{sample-base}

\end{document}